\documentclass[conference]{IEEEtran}
\IEEEoverridecommandlockouts

\usepackage{cite}
\usepackage{amsmath,amssymb,amsfonts}
\usepackage{textcomp}
\usepackage{xcolor}
\usepackage{url}
\usepackage{graphicx}
\usepackage{booktabs}
\usepackage{multirow}
\usepackage{placeins}
\usepackage{float}
\usepackage{cuted}
\usepackage{capt-of}
\usepackage{xspace}

\newcommand{\methodfullname}{\textsc{MemoryLACE}\xspace}
\newcommand{\methodname}{\textsc{MemLACE}\xspace}

\begin{document}

\title{\methodfullname: Memory Lifecycle-Aware Consolidation and Evidence Retrieval}


\author{
\IEEEauthorblockN{
Meriem Yacoubi\IEEEauthorrefmark{1}\IEEEauthorrefmark{2},
Pia Schmidt\IEEEauthorrefmark{2},
Nenad Petrovic\IEEEauthorrefmark{1},
Ahmed Frikha\IEEEauthorrefmark{3},
Martin Kirchhoff\IEEEauthorrefmark{2},
Alois Knoll\IEEEauthorrefmark{1}
\vspace{0.2cm}
}

\IEEEauthorblockA{\IEEEauthorrefmark{1}
\textit{Chair of Robotics, Artificial Intelligence and Real-Time Systems, Technical University of Munich} \\
}

\IEEEauthorblockA{\IEEEauthorrefmark{2}
\textit{inovex GmbH}
}

\IEEEauthorblockA{\IEEEauthorrefmark{3}
\textit{Cerebras Systems Inc.}
}

Email: \{meriem.yacoubi, nenad.petrovic, k\}@tum.de,\\
\{pia.schmidt, martin.kirchhoff\}@inovex.de,
ahmed.frikha@cerebras.net
}
\maketitle

\begin{abstract}
Long-term LLM agents must preserve information across interactions while distinguishing repeated evidence, historical states, updates, and unresolved contradictions. Existing textual memory systems retrieve semantically relevant memories efficiently but often leave these relationships implicit, whereas richer structured approaches model them through global graphs, hierarchical abstractions, or reflection at greater complexity. We introduce \methodfullname{} (\methodname), a lightweight memory framework that explicitly models the lifecycle of textual evidence through sparse merge, supersession, and contradiction relations while preserving atomic natural-language memories and their provenance. Rather than retrieving memories independently, \methodname reconstructs relation-aware evidence units that expose current, historical, supporting, and conflicting evidence for downstream reasoning. Across BEAM and StructMemEval, using open-weight and proprietary LLM backbones, \methodname achieves the highest overall performance in same-backbone comparisons while reducing end-to-end runtime on BEAM by 66.6\% relative to Hindsight, the strongest reported reflective-memory baseline. Ablation studies identify lifecycle expansion and temporal awareness as the principal contributors to these gains. Together, the results demonstrate that explicitly modeling the local lifecycle of textual evidence is sufficient to substantially improve long-term memory reasoning without requiring comprehensive knowledge graphs or global reflection.
\end{abstract}

\begin{IEEEkeywords}
Long-term memory, LLM agents, evidence lifecycle, memory consolidation, contradiction handling
\end{IEEEkeywords}

\section{Introduction}

Large Language Models (LLMs) increasingly support agents that interact with users across tasks and sessions. Beyond a single context window, these agents require memory mechanisms that preserve preferences, plans, corrections, constraints, task progress, and previously established facts~\cite{hu2025memorysurvey}. External memory can ground generation~\cite{lewis2020rag}, support persistent behavior~\cite{park2023generative}, and extend long-term interaction~\cite{packer2023memgpt}. However, reliable memory requires more than retrieving relevant text: new evidence may repeat, replace, or contradict stored information. Existing approaches expose a practical trade-off. Compact textual systems offer efficient storage and retrieval~\cite{simplemem2026,mem0}, but often leave relationships between memories implicit. Structured, hierarchical, and reflective architectures represent these relationships explicitly~\cite{amem,zep2025,timem,himem2026}, but may require broader graph construction, abstraction, or reflection. This motivates a focused question: how can an agent preserve changes in textual evidence and retrieve dependent memories jointly without constructing a comprehensive global structure?

We introduce \methodfullname{} (\methodname), a lightweight middle ground between flat textual and comprehensive structured memory. \methodname preserves atomic natural-language memories and their provenance while connecting related entries through sparse local relations for merge, supersession, and contradiction. During writing, bounded candidate comparison and local batch evolution establish and repair these relations. During retrieval, active memories provide initial anchors, lifecycle expansion recovers connected historical or conflicting evidence, and relation-aware evidence units are reranked and packed under the context budget. The answer model therefore receives current, historical, supporting, and conflicting evidence in an interpretable structure rather than as independent snippets.

Our contributions are threefold. First, we introduce an explicit lifecycle representation and consolidation framework that distinguishes repeated evidence, ordered state changes, and unresolved contradictions while preserving atomic natural-language memories and their provenance. Second, we propose a relation-aware retrieval mechanism that reconstructs lifecycle-aware evidence units by expanding active memories through lifecycle relations, enabling the language model to jointly interpret current, historical, supporting, and conflicting evidence rather than isolated memory snippets. Third, through comprehensive evaluation on \emph{BEAM} and \emph{StructMemEval} using both open-weight and proprietary LLM backbones, together with ablation studies, we demonstrate that explicitly modeling only local evidence lifecycle structure is sufficient to substantially improve long-term memory reasoning, achieving the strongest overall performance in same-backbone comparisons without requiring a comprehensive knowledge graph or global reflection stage.

\section{Related Work}

Recent surveys characterize agent memory through how information is represented, used, evolved, and retrieved over time~\cite{hu2025memorysurvey}. Within textual memory, Mem0 extracts salient conversational information, reconciles it with stored entries, and retrieves the resulting memory objects across sessions~\cite{mem0}, while SimpleMem emphasizes semantic compression, recursive consolidation, and query-aware retrieval~\cite{simplemem2026}. SeCom studies memory-unit granularity and constructs compressed memory segments organized around coherent topics~\cite{secom2025}, and MemInsight augments stored interactions to improve semantic representation and retrieval~\cite{meminsight2025}. EMem represents dialogue as structured event-level propositions with normalized entities, temporal cues, and source attribution, optionally connecting them through a lightweight heterogeneous graph~\cite{emem2025}. These systems show that compact textual or event-level memories can provide strong long-horizon recall, but generally do not represent repetition, supersession, and unresolved contradiction as explicit lifecycle relations.

A-MEM, Zep, Theanine, TiMem, and HiMem provide closely related approaches to evolving and temporally organized memory. A-MEM constructs structured notes and allows new evidence to trigger the evolution of linked representations~\cite{amem}. Zep uses a temporally aware knowledge graph~\cite{zep2025}, Theanine links memories through temporal and causal timelines~\cite{theanine2025}, and TiMem and HiMem consolidate observations into hierarchical episodic and semantic representations~\cite{timem,himem2026}. These systems demonstrate the value of temporal continuity, but their representations model broader semantic, causal, or hierarchical structure. \methodname instead creates only local merge, supersession, and contradiction relations when they alter the interpretation of stored evidence.

Reflective Memory Management, Hindsight, and MemOS address memory evolution through richer maintenance mechanisms. Reflective Memory Management maintains both prospective and retrospective user representations~\cite{reflective2025}, while Hindsight separates retained observations, recalled information, and reflective interpretations across distinct memory networks~\cite{hindsight}. MemOS treats memory as a governed system resource, supporting provenance, versioning, composition, migration, and evolution across heterogeneous memory forms~\cite{memos2025}. These approaches reinforce the importance of preserving evidence while allowing its interpretation and operational status to evolve. Rather than introducing dedicated reflection modules or a memory operating layer, \methodname limits maintenance to bounded candidate comparison, sparse lifecycle relations, and relation-aware retrieval.

Benchmarks for long-term memory have moved beyond simple conversational recall. LoCoMo and LongMemEval probe multi-session question answering, temporal reasoning, knowledge updates, and abstention~\cite{locomo2024,longmemeval2025}, while MemBench and Evo-Memory add efficiency, capacity, and continuous memory evolution as evaluation axes~\cite{membench2025,evomemory2025}. We evaluate on \emph{BEAM}~\cite{beam2025} and \emph{StructMemEval}~\cite{structmemeval}, which test complementary abilities: \emph{BEAM} measures long-context conversational memory across ten capabilities, whereas \emph{StructMemEval} isolates whether accumulated evidence can be organized to support structured reasoning. Together they cover both the interpretation of evolving evidence and its structured use, the two settings our method targets.

\section{\methodfullname{} Architecture}

\methodname treats long-term memory as evolving evidence rather than a flat collection of summaries. The architecture comprises a memory representation and two stages. The representation models atomic textual memories with temporal metadata, provenance, activity state, and sparse lifecycle relations. The first stage, \emph{memory construction and consolidation}, converts dialogue into standalone memories, stores them, and determines how new evidence relates to existing memories. The second stage, \emph{evidence retrieval and context construction}, retrieves active anchors, i.e., active memories selected as relevant to the query, expands their local relations, and organizes connected entries before answer generation. This design enables the system to retrieve relevant evidence while preserving its evolution over time, without requiring a comprehensive, globally normalized knowledge graph.

\subsection{Memory Representation}
Each stored memory is represented as
\[
m_i=(x_i,\tau_i,z_i,p_i,a_i,R_i),
\]
where \(x_i\) is a standalone natural-language restatement of one atomic piece of explicit dialogue evidence. The restatement preserves the original meaning and relevant details while making the memory independently understandable outside the original conversation context. The field \(\tau_i\) records evidence-grounded temporal metadata, such as an explicitly stated event time or validity period, while \(z_i\) contains lightweight metadata such as entities, keywords, topics, and locations. The provenance field \(p_i\) records the identifiers of the supporting dialogue turns, thereby preserving the order and origin of the evidence; \(a_i\) indicates whether the memory is active; and \(R_i\) denotes the set of lifecycle-relation references associated with the memory, including zero or more merge, supersession, and contradiction relations. Here, the lifecycle of a memory is how its status evolves as new evidence arrives: whether it remains active and how it relates to repeated, superseded, superseding, or conflicting memories. Supersession is represented in both directions: newer memories point to the entries they replace, and older memories point to their successors.

The complete memory store forms a sparse relation graph \(G=(M,E)\), where \(M\) is the set of textual memory nodes and \(E\) comprises directed supersession edges and symmetric merge and contradiction links. Unlike a general knowledge graph, which may encode entity, taxonomic, causal, or semantic relations, \methodname stores only relations needed to interpret stored evidence. Provenance identifies a memory's source, whereas lifecycle relations describe its connections to other memories. This preserves the flexibility of natural-language memory while making its evolution inspectable.

\subsection{Stage I: Memory Construction and Consolidation}
\begin{figure*}[t]
\centering
\includegraphics[width=\linewidth]{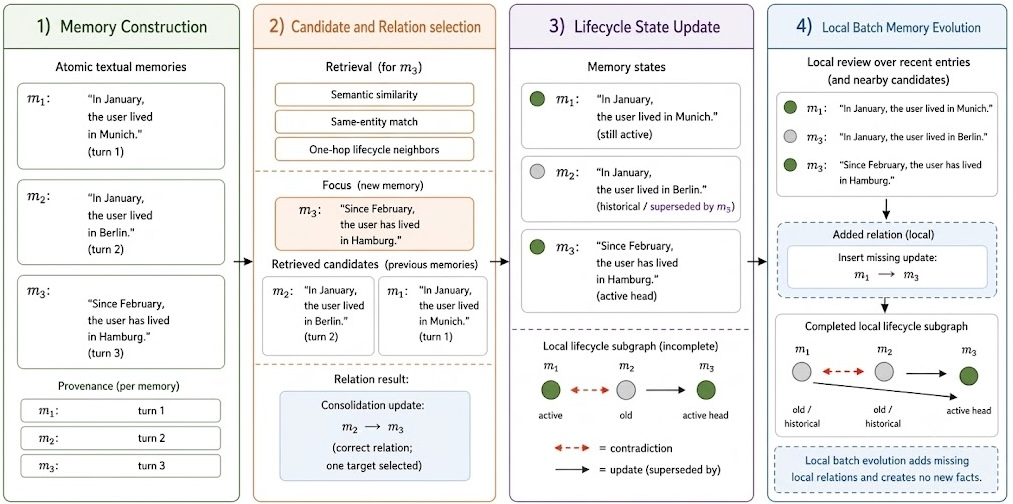}
\caption{Memory construction and consolidation in \methodname. Atomic provenance-bearing memories undergo bounded relation selection, lifecycle-state updates, and local batch repair of missed relations.}
\label{fig:memory-construction}
\end{figure*}

Figure~\ref{fig:memory-construction} illustrates Stage~I. The process follows the lightweight sliding-window memory construction of SimpleMem~\cite{simplemem2026}: overlapping dialogue windows are converted into standalone atomic memories, and duplicate entries introduced by the overlap are consolidated before lifecycle processing. \methodname extends this process with lifecycle-aware consolidation for evidence that evolves over time.

\subsubsection{Memory Construction}

For a dialogue window \(D_t\), the memory builder produces
\[
B(D_t)\rightarrow\{m_1,m_2,\ldots,m_k\}.
\]
Each atomic entry captures a single explicit fact, preference, plan, correction, decision, event, or state, together with source-turn provenance that links it to the supporting dialogue evidence. Names, dates, quantities, negations, and speaker attribution are preserved as stated, and no unstated information is inferred; mentioning a trip to Berlin, for instance, does not imply residence there. This atomic granularity is important because an update or contradiction typically affects only a single claim within a broader dialogue segment, allowing consolidation to act on precisely the affected memory. Since overlapping windows may extract the same claim from the same turns more than once, such duplicate extractions are collapsed and their provenance combined, whereas repetitions from distinct turns remain separate for possible merge linking.

\subsubsection{Candidate and Relation Selection}

Memories are consolidated in chronological order so that later evidence cannot be treated as preceding the state it modifies. For each memory \(m_{\mathrm{new}}\), \methodname first retrieves related memories through semantic similarity and same-entity matching, then expands this set with their one-hop lifecycle neighbors to form the candidate set \(C(m_{\mathrm{new}})\). These neighbors are memories connected through merge, supersession, or contradiction relations. A cross-encoder reranks the candidates, preserving lifecycle-linked memories and prioritizing the highest-ranked remaining candidates. Reranking determines only which memories are considered for comparison; it does not determine their relation to \(m_{\mathrm{new}}\).

The consolidation module subsequently selects exactly one strategy \(r\) and, when applicable, one target memory. Restricting each step to a single relation--target pair lets the model focus on one decision at a time, which makes the strategy decision more reliable than jointly assigning relations to several candidates:
\[
\mathcal{R}=
\{\mathrm{merge},\mathrm{update},
\mathrm{contradict},\mathrm{no\_action}\},
\]
\[
g(m_{\mathrm{new}},C(m_{\mathrm{new}}))
\rightarrow(r,c^{\star}),
\qquad
r\in\mathcal{R}.
\]
Here, \(g\) denotes the consolidation decision function, and \(c^{\star}\in C(m_{\mathrm{new}})\) denotes the selected target memory, or null when no target is required. \emph{Merge} is applied to duplicate evidence about the same fact, \emph{update} to a time-ordered change of the same state, \emph{contradict} to incompatible claims with the same scope, and \emph{no action} to independent memories. Because retrieval identifies only potentially related candidates, the consolidation model assesses their subject, attribute, claim, and temporal scope before assigning a relation, preventing topical similarity alone from altering a memory's lifecycle state. Relations to additional memories are recovered by the subsequent local batch step rather than forced into this single decision.

\subsubsection{Lifecycle State Update}

Under the default configuration, lifecycle decisions modify memory activity, i.e., whether a memory belongs to the active set \(A\), and connectivity, i.e., its lifecycle relations to other memories, without deleting source evidence. Let
\[
A=\{m_i\in M\mid a_i=1\}
\]
denote the active subset of the complete memory store \(M\). When \(m_n\) updates \(m_o\), the older entry becomes inactive and the newer entry becomes the active chain head:
\[
a_o\leftarrow 0,\qquad a_n\leftarrow 1,
\]
\[
m_n\xrightarrow{\mathrm{supersedes}}m_o,
\qquad
m_o\xrightarrow{\mathrm{superseded\_by}}m_n.
\]
Contradictory memories remain active because the available evidence does not resolve their conflict, and they are linked symmetrically. Memories assigned no action remain independently retrievable. For merge, we explore two consolidation policies. Under the \emph{default linked-merge} policy, memories assigned the merge relation retain their original text and provenance while receiving bidirectional merge references, so that each original piece of evidence remains separately retrievable. In contrast, the \emph{compact-merge} policy replaces the merged entries with one synthesized active memory, producing a smaller memory store at the cost of discarding the original individual entries. Unless otherwise stated, \methodname uses the default linked-merge policy. Independently, superseded states are excluded from ordinary active retrieval while historical, merge-linked, and conflicting evidence remains stored.

\subsubsection{Local Batch Memory Evolution}

Each memory is consolidated through a single lifecycle decision, keeping every per-memory step bounded and inspectable while reducing sensitivity to decision order. After a new group of memories has been added, \methodname complements this with a bounded local evolution step that considers the group jointly. It reviews the newly added memories together with their Stage~I consolidation candidates and establishes any further update, contradiction, or merge links that become apparent only when the group is viewed together, such as a memory relating to more than one earlier entry. This step acts only on lifecycle structure, updating relation links and corresponding activity states without rewriting memory text or adding factual content. Coordinating single-decision consolidation with group-level evolution therefore repairs additional lifecycle relations while preserving the bounded, local decision process.
\subsection{Stage II: Evidence Retrieval and Context Construction}

\begin{figure*}[t]
\centering
\includegraphics[width=\linewidth]{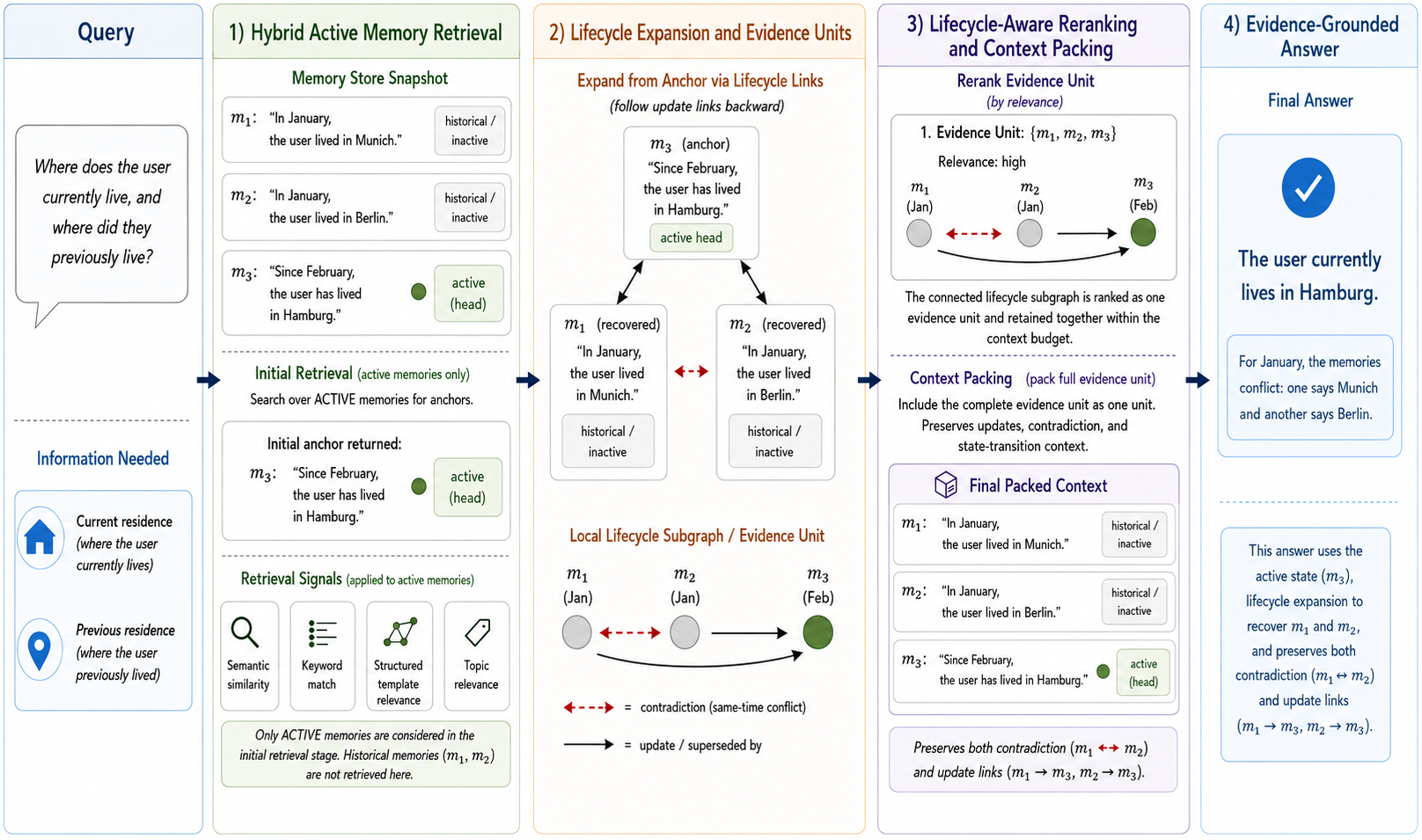}
\caption{Evidence retrieval and context construction in \methodname. Active anchors are expanded through lifecycle relations, grouped into evidence units, reranked, and packed for evidence-grounded generation.}
\label{fig:read-phase}
\end{figure*}

Figure~\ref{fig:read-phase} summarizes the query-time pipeline.

\subsubsection{Hybrid Active-Memory Retrieval}

At query time, we use semantic similarity, keyword match, structured template relevance, and topic relevance to produce complementary ranked lists over active memories. The resulting rankings are combined using Reciprocal Rank Fusion (RRF)~\cite{cormack2009rrf}, which does not require their raw scores to be directly comparable:
\[
s_{\mathrm{RRF}}(m)=
\sum_{c\in\mathcal{C}}
\frac{1}{k+\operatorname{rank}_{c}(m)},
\]
where \(\mathcal{C}\) denotes the retrieval channels and \(k\) is a rank-smoothing constant. Restricting the initial search to active memories reduces competition from superseded states. Inactive memories remain stored and can be recovered via retrieved anchors' lifecycle relations when historical evidence is required. Formally, hybrid retrieval returns the set of active anchors
\[
A_q = R_{\mathrm{hybrid}}(q, A),
\]
where \(A\) is the active-memory set and \(R_{\mathrm{hybrid}}\) fuses the retrieval channels above through RRF.

\subsubsection{Lifecycle Expansion and Evidence Units}
Hybrid retrieval returns relevant active anchors \(A_q\), which alone may not contain all evidence required to interpret how the corresponding information evolved over time. \methodname therefore expands each anchor through bounded lifecycle relations,

\[
E_q = A_q \cup L(A_q),
\]
where \(L(A_q)\) follows supersession lineages to recover prior states and one-hop contradiction relations to surface directly conflicting evidence. The expansion remains local rather than traversing the full memory graph, limiting additional context introduced for each query while still recovering the historical and conflicting evidence that a query may depend on.

The resulting memories are grouped into evidence units. A unit can represent a standalone memory or a connected component formed by supersession, contradiction, and merge links among memories in the retrieved pool. Supersession edges arrange each component from earlier to newer states so that both historical evidence and the active state remain visible. Unresolved contradictory claims are retained together rather than prematurely resolving one in favor of another. Grouping before reranking allows memories whose interpretation depends on their relations to be evaluated as a coherent unit instead of competing as independent snippets.

\subsubsection{Chain-Aware Reranking and Context Packing}

A reranker scores the resulting lifecycle-aware evidence units against the user query, after which the highest-scoring units are selected for answer generation. Because evidence units are scored as whole groups, relation-connected memories are evaluated jointly during reranking rather than competing as independent snippets. The highest-ranked units are then selected subject to the context budget. Update chains preserve their older-to-newer order, and complete evidence units are retained whenever the budget permits. When a unit would exceed the remaining budget, it is included only in part, keeping its higher-priority memories and dropping the rest.

\subsubsection{Evidence-Grounded Answer Generation}

During answer generation, the packed context is presented in lifecycle order, distinguishing current, superseded, and conflicting memories to the LLM. The model can therefore prioritize active evidence for current-state questions, recover earlier states for historical or sequential questions, and retain competing claims when contradictions are unresolved, receiving relation-connected evidence with its temporal structure preserved rather than inferring relations from unordered snippets.

\section{Experimental Setup}
We evaluate \methodname on two long-term memory benchmarks. \emph{BEAM}~\cite{beam2025} is the main benchmark for conversational long-term memory, while \emph{StructMemEval}~\cite{structmemeval} assesses whether evidence can support structured reasoning.

\subsection{Benchmarks}
\emph{BEAM} evaluates ten different memory capabilities over long conversational histories~\cite{beam2025}. We use its 100K-token subset, which contains 20 conversations and 400 questions. \emph{StructMemEval} evaluates state tracking, tree-based reasoning, count-based reasoning, and recommendation over persistently stored information~\cite{structmemeval}. We use its 51-scenario main set, comprising the longest cases in each task family.

\subsection{Experimental Configuration}
For a fair comparison, we align \methodname with the model and evaluation settings of the corresponding baselines. On BEAM, the evaluated backbones are Qwen3.5-4B and Qwen3.5-9B (both FP8)~\cite{qwen35}. On StructMemEval, \methodname and Mem-Agent are evaluated with GPT-5.5~\cite{gpt55}. For each benchmark, the backbone is used for memory construction, consolidation, batch evolution, query analysis, and answer generation. Generated answers are scored by GPT-4o mini~\cite{gpt4omini} as the fixed LLM judge following each benchmark's evaluation protocol. Reasoning mode was disabled for all Qwen3.5 runs, and the generation temperature was set to \(0.0\).

Memory construction uses 20-turn windows with a 2-turn overlap. Consolidation retrieves 12 semantic and 8 entity candidates under a total candidate budget of 15, reranks the five highest-scoring candidates, and reviews at most 24 entries during batch evolution. Query-time retrieval uses active-only anchors and retrieves up to 40 semantic, 30 keyword, 30 structured, and 20 topic candidates, which are fused using RRF with \(k=60\). Supersession-lineage expansion is bounded at 80 nodes, contradiction expansion at four memories, and the final context at 64 memories. Semantic retrieval uses Qwen/Qwen3-Embedding-0.6B~\cite{qwen3embedding}, while candidate and evidence-unit reranking use cross-encoder/ms-marco-MiniLM-L6-v2~\cite{minilmreranker}.

\section{Results and Analysis}

\begin{table*}[!t]
\centering
\small
\setlength{\tabcolsep}{2pt}
\caption{Category-level results on the \emph{BEAM} 100K subset (in \%)}
\label{tab:beam-category-results}
\begin{tabular}{l|l|cccccccccc|c}
\toprule
Method
& Backbone
& Abst.
& Contr.
& Event
& Info.
& Instr.
& Update
& Multi.
& Pref.
& Summ.
& Temp.
& Overall \\
\midrule
SimpleMem
& Qwen3.5-4B
& 38.8 & 7.2 & 21.9 & 56.0 & 29.4
& 31.9 & 26.9 & 53.1 & 7.7 & \textbf{32.5} & 30.5 \\

LIGHT
& Qwen3.5-4B
& 62.5 & 13.8 & 2.6 & \textbf{57.3} & 45.0
& \textbf{35.6} & 38.7 & 63.3 & 29.8 & 7.5 & 35.6 \\

Hindsight
& Qwen3.5-4B
& 50.0 & 6.6 & 18.7 & 50.0 & 59.4
& 30.0 & \textbf{40.3} & \textbf{85.0} & \textbf{33.4}
& 28.1 & 40.1 \\

\methodname
& Qwen3.5-4B
& \textbf{88.8} & \textbf{36.6} & \textbf{29.3} & 52.9
& \textbf{61.9} & 25.0 & 33.8 & 81.3 & 27.8 & 16.9
& \textbf{45.4} \\
\midrule
SimpleMem
& Qwen3.5-9B
& 37.5 & 10.0 & 21.2 & 51.2 & 30.0
& 36.9 & 33.7 & 48.8 & 7.8 & 30.0 & 30.7 \\

LIGHT
& Qwen3.5-9B
& \textbf{75.0} & 13.8 & 2.3 & \textbf{67.7} & 51.2
& \textbf{52.5} & 38.9 & 71.9 & 36.0 & 11.9 & 42.1 \\

Hindsight
& Qwen3.5-9B
& 63.7 & \textbf{40.9} & 20.8 & 61.1 & 68.1
& 51.2 & 50.3 & 71.2 & \textbf{40.4} & \textbf{35.0}
& 50.3 \\

\methodname
& Qwen3.5-9B
& 71.3 & 40.3 & \textbf{25.6} & 55.9 & \textbf{72.5}
& 51.9 & \textbf{55.3} & \textbf{84.4} & 35.5 & 26.3
& \textbf{51.9} \\
\bottomrule
\end{tabular}
\end{table*}

We evaluate \methodname on the \emph{BEAM} 100K subset and the \emph{StructMemEval} main set. On \emph{BEAM}, we compare against LIGHT~\cite{beam2025}, reported by the benchmark as its highest-performing memory architecture, Hindsight~\cite{hindsight}, which uses a retain--recall--reflect pipeline with global reflection, and SimpleMem~\cite{simplemem2026}, a compact textual-memory system that shares our standalone construction scheme, isolating the contribution of the lifecycle representation and retrieval. We use Qwen3.5 models as backbones for our \emph{BEAM} evaluations. On \emph{StructMemEval}, we compare against Mem-Agent~\cite{memagent2025}, a markdown-based memory agent, under the same GPT-5.5 backbone.

\subsection{\emph{BEAM} Category-Level Comparison}

Table~\ref{tab:beam-category-results} reports our results across the ten \emph{BEAM} memory capabilities. The column abbreviations denote abstention (Abst.), contradiction resolution (Contr.), event ordering (Event), information extraction (Info.), instruction following (Instr.), knowledge update (Update), multi-hop reasoning (Multi.), preference following (Pref.), summarization (Summ.), and temporal reasoning (Temp.).

We find that \methodname substantially outperforms all baselines on the overall average at both model sizes. With Qwen3.5-4B as backbone, \methodname achieves the best results among the 4B systems in abstention, contradiction resolution, event ordering, and instruction following, indicating that the benefit of lifecycle structure already appears at smaller model scale without requiring a larger LLM backbone to compensate for disorganized context.

With the Qwen3.5-9B backbone, \methodname outperforms the strongest baseline, Hindsight, in six of ten categories. The gains are not spread uniformly: they concentrate in preference following, abstention, multi-hop reasoning, event ordering, and instruction following, precisely the categories in which the correct answer depends on considering several related memories together rather than in isolation. This is consistent with lifecycle-aware retrieval, which delivers the active state alongside the historical and supporting evidence connected to it. Knowledge update improves only marginally, which is expected: a single active-memory lookup already suffices and does not benefit from adding and grouping related entries.

Although \methodname achieves the highest 9B score, Hindsight retains an advantage of \(8.7\) percentage points in temporal reasoning, \(5.2\) percentage points in information extraction, and \(4.9\) percentage points in summarization. This pattern reflects the different design priorities of the two systems. \methodname deliberately uses atomic memories, sparse local relations, and bounded evidence packing to resolve evolving states efficiently, rather than reconstructing the complete dialogue or maintaining a globally reflective representation as Hindsight does. While this trade-off yields substantially higher efficiency, as shown in the next section, it is less suited to exhaustive summarization and explicit temporal-interval calculation.

\subsubsection{Runtime Comparison}
In this section, we compare the runtime needed by \methodname to that of the highest-performing baseline method, Hindsight, using Qwen3.5-9B as backbone. Table~\ref{tab:runtime} reports the phase-level breakdown under the same evaluation settings, i.e., the \emph{BEAM} 100K subset, vLLM serving environment, parallel-processing configuration, and one NVIDIA A40 virtual GPU profile with 24\,GB of allocated memory.

\begin{table}[H]
\centering
\small
\setlength{\tabcolsep}{4pt}
\caption{Runtime breakdown on the \emph{BEAM} 100K subset with Qwen3.5-9B.}
\label{tab:runtime}
\begin{tabular}{lccc}
\toprule
Phase & \methodname & Hindsight & Speedup \\
\midrule
Memory construction & 6\,h\,06\,min & 18\,h\,27\,min & $3.02\times$ \\
Answer generation   & 1\,h\,25\,min & 4\,h\,03\,min  & $2.86\times$ \\
\midrule
Total               & 7\,h\,31\,min & 22\,h\,30\,min & $2.99\times$ \\
\bottomrule
\end{tabular}
\end{table}

\methodname reduces the total system runtime by 66.6\%, a $2.99\times$ speedup. The largest saving occurs in memory construction ($3.02\times$), the phase in which \methodname employs bounded consolidation and local batch repair instead of the expensive global reflection stage used by Hindsight.

\subsection{\emph{StructMemEval} Results}
The results on the \emph{StructMemEval} benchmark are presented in Table~\ref{tab:struct-gpt55}. We compare \methodname with Mem-Agent~\cite{memagent2025}, the strongest among the memory systems evaluated on \emph{StructMemEval}~\cite{structmemeval}, using GPT-5.5 as the backbone LLM. A scenario counts as solved when at least half of its questions are judged correct, and the table reports the fraction of solved scenarios per subtask together with their macro-average.

\begin{table}[H]
\centering
\small
\setlength{\tabcolsep}{2pt}
\caption{\emph{StructMemEval} results using GPT-5.5 as the LLM (in \%).}
\label{tab:struct-gpt55}
\begin{tabular}{l|cccc|c}
\toprule
Method
& State
& Tree
& Count
& Recsys
& Overall \\
\midrule
Mem-Agent
& 57.00
& 50.00
& 0.00
& \textbf{33.00}
& 35.00 \\
\methodname
& \textbf{100.00}
& \textbf{100.00}
& 0.00
& 8.33
& \textbf{52.08} \\
\bottomrule
\end{tabular}
\end{table}

\begin{table*}[!t]
\centering
\small
\setlength{\tabcolsep}{3pt}
\caption{Ablation study results using Qwen3.5-9B on \emph{BEAM} 100K (in \%), including the absolute change from the full \methodname configuration ($\Delta$). The combined variant uses compact merge and disables both rerankers, batch evolution, temporal awareness, and lifecycle expansion.}
\label{tab:ablation-detailed}
\begin{tabular}{l|cccccccccc|cc}
\toprule
Variant
& Abst.
& Contr.
& Event
& Info.
& Instr.
& Update
& Multi.
& Pref.
& Summ.
& Temp.
& Overall
& $\Delta$ \\
\midrule
Full \methodname
& 71.3 & 40.3 & 25.6 & 55.9 & 72.5
& 51.9 & 55.3 & 84.4 & 35.5 & 26.3 & \textbf{51.9} & -- \\
\midrule
Without both rerankers
& 70.0 & 39.1 & 23.4 & 57.9 & 63.8
& 56.3 & 50.8 & 80.6 & 37.3 & 27.5 & 50.7 & -1.23 \\

Compact merge
& 71.3 & 44.1 & 24.7 & 54.8 & 71.3
& 53.8 & 43.0 & 83.8 & 34.4 & 23.8 & 50.5 & -1.43 \\

Without batch evolution
& 72.5 & 36.9 & 24.1 & 56.5 & 60.6
& 55.0 & 48.2 & 83.3 & 38.1 & 26.9 & 50.2 & -1.68 \\

Without temporal awareness
& 72.5 & 26.6 & 23.8 & 54.4 & 67.5
& 48.8 & 46.9 & 75.6 & 37.0 & 16.3 & 46.9 & -4.97 \\

Without lifecycle expansion
& 73.8 & 40.3 & 23.4 & 47.8 & 69.4
& 35.0 & 42.3 & 85.8 & 35.9 & 13.8 & 46.7 & -5.16 \\
\midrule
Combined removal
& 67.5 & 23.1 & 25.2 & 48.5 & 69.4
& 36.3 & 37.6 & 78.8 & 35.9 & 11.3 & 43.3 & -8.55 \\
\bottomrule
\end{tabular}
\end{table*}

\methodname outperforms Mem-Agent by more than 17 percentage points. \methodname solves all 14 state-tracking and all 10 tree-based scenarios, against 57.00\% and 50.00\% for Mem-Agent. The state-tracking result benefits directly from the memory representation we propose: supersession links preserve each state and active-head management resolves the current memory, thereby enabling correct answers to both current-state and historical questions. The tree-based result is more notable, as it requires reconstructing hierarchical relations, a setting often associated with graph-based memory rather than flat textual stores. \methodname reconstructs the queried hierarchy from atomic relational memories, structured retrieval, and evidence-unit grouping, supplying the answer model with the locally relevant dependencies rather than a comprehensive hierarchical knowledge graph. Perfect accuracy on both families lifts the macro-average to 52.08\%, and the gap over Mem-Agent is driven entirely by these two structured, evolving-state task types.

The counting and recommendation results identify a targeted extension opportunity for the framework. Neither method solves the count-based scenarios. In recommendation, \methodname answers \(32.2\%\) of the individual questions correctly, as computed from the question-level judge outputs, while reaching \(8.33\%\) at the scenario level under the benchmark's \(50\%\) correctness threshold. These tasks depend on global numerical accumulation or ranking across many distributed preference signals, whereas \methodname is designed to preserve validity, provenance, and dependencies among evolving pieces of evidence. Dedicated aggregation operators, e.g., invoking a calculator or code-execution tool, could complement the evidence units produced by \methodname without requiring a more expensive global memory architecture.

\section{Ablation Study}
\label{sec:ablation}

We conduct an ablation study to gain further insight into the importance and empirical impact of each of the \methodname components. We isolate the contribution of each component by evaluating variants of \methodname on the complete \emph{BEAM} 100K subset using Qwen3.5-9B. Each run changes exactly one component, except the \emph{combined removal} variant, which uses compact merge and disables both rerankers, batch evolution, temporal awareness, and lifecycle expansion. Table~\ref{tab:ablation-detailed} reports category-level results for every variant.

Lifecycle expansion and temporal awareness account for the two largest reductions, but their effects fall on distinct capabilities, which indicates that they address different failure modes. On the one hand, removing lifecycle expansion (\(-5.16\)) leaves active-memory retrieval intact but stops it from traversing supersession and contradiction links, and the loss concentrates on categories that require earlier and current states together: knowledge update ($51.9\rightarrow35.0$), multi-hop reasoning ($55.3\rightarrow42.3$), and temporal reasoning ($26.3\rightarrow13.8$). Contradiction resolution is unchanged at the reported precision ($40.3$ in both), indicating that the relevant conflicting memories were often retrieved independently as active anchors even without contradiction-link expansion. On the other hand, removing temporal awareness (\(-4.97\)) degrades contradiction resolution ($40.3\rightarrow26.6$), preference following ($84.4\rightarrow75.6$), and temporal reasoning ($26.3\rightarrow16.3$). Temporal awareness here denotes \methodname's use of evidence-grounded temporal metadata (\(\tau_i\)), such as event times and validity periods, when consolidating and ordering memories, so that an ordered change over time is distinguished from a genuine conflict; without it, a valid update is more easily misread as a contradiction. The two components are therefore not interchangeable: connectivity determines which related memories are retrieved, whereas temporal grounding determines whether a difference between them is read as a genuine contradiction or an ordered change.

The remaining components lead to smaller accuracy reductions. The impacted categories clarify the contribution of each component. Removing batch evolution (\(-1.68\)) affects instruction following ($72.5\rightarrow60.6$) and multi-hop reasoning ($55.3\rightarrow48.2$) most, consistent with its role of recovering relations that span entries which were separately handled by the single-decision consolidation. Removing both rerankers (\(-1.23\)) produces the smallest performance drop, although instruction following decreases from $72.5$ to $63.8$, consistent with a component that reweights already-retrieved evidence rather than changing which evidence or relations exist. The merge comparison is the most diagnostic: the compact merge policy, which replaces the default linked-merge entries with a single synthesized memory, improves contradiction resolution ($40.3\rightarrow44.1$) but substantially reduces multi-hop reasoning ($55.3\rightarrow43.0$) for a net drop of 1.43 percentage points. Consolidating evidence into one entry thus helps when a single reconciled statement is sufficient, but discards the separate provenance and phrasing that multi-hop questions rely on. The default linked-merge policy applied in \methodname strikes a better trade-off.

The combined removal drops the overall score to $43.3\%$, a larger reduction than any single ablation. We note that this combined reduction is smaller than the sum of the individual drops, indicating that the components overlap partially while still contributing separable effects. The degradation is concentrated on evolving and relation-dependent categories, i.e., contradiction resolution ($23.1$), knowledge update ($36.3$), multi-hop reasoning ($37.6$), and temporal reasoning ($11.3$), whereas event ordering ($25.2$) and summarization ($35.9$) remain close to the full configuration.

\section{Conclusion}

We presented \methodfullname{} (\methodname), a long-term memory framework that represents the evolution of textual evidence through sparse lifecycle relations connecting repeated, superseded, and contradictory memories while preserving their original provenance. By retrieving and reasoning over relation-aware evidence units rather than isolated memories, \methodname enables language models to interpret current, historical, supporting, and conflicting evidence within a unified retrieval pipeline. Across BEAM and StructMemEval, \methodname achieves the strongest overall performance in same-backbone comparisons while substantially reducing runtime relative to Hindsight, the strongest reported reflective-memory baseline. Ablation studies further show that lifecycle expansion and temporal awareness account for most of these gains, providing empirical evidence that explicitly modeling the local lifecycle of textual evidence is sufficient to recover much of the reasoning benefit typically associated with substantially heavier memory architectures. These findings suggest that, in the evaluated settings, the principal challenge for long-term memory is not richer global structure but preserving how evidence evolves over time. Future work can therefore extend this lifecycle-aware interface with specialized operators for aggregation, counting, and numerical reasoning while retaining the same lightweight memory representation.

\clearpage

\end{document}